\documentclass[journal,10pt]{IEEEtran}
\usepackage{multirow}
\usepackage{tabu}
\usepackage{makecell}
\usepackage{color}
\usepackage[table]{xcolor}
\usepackage{graphicx}
\usepackage{subfigure}
\usepackage{bm}
\usepackage{balance}
\usepackage{amsmath,amssymb,amsfonts}

\usepackage{multirow}

\usepackage{enumitem}

\usepackage[ruled,linesnumbered]{algorithm2e}

\SetAlCapNameFnt{\scriptsize}
\SetAlCapFnt{\scriptsize}

\SetCommentSty{mycommfont}

\begin{document}
\title{A Lightweight Concept Drift Detection and Adaptation Framework for IoT Data Streams}

\author{\IEEEauthorblockN{Li Yang and Abdallah Shami}\\
\IEEEauthorblockA{
Western University, London, Ontario, Canada \\
e-mails: \{lyang339, abdallah.shami\}@uwo.ca}}

% no key words

\markboth{Accepted and to appear in IEEE Internet of Things Magazine}
{}

\maketitle

\begin{abstract}
In recent years, with the increasing popularity of “Smart Technology”, the number of Internet of Things (IoT) devices and systems have surged significantly. Various IoT services and functionalities are based on the analytics of IoT streaming data. However, IoT data analytics faces concept drift challenges due to the dynamic nature of IoT systems and the ever-changing patterns of IoT data streams. In this article, we propose an adaptive IoT streaming data analytics framework for anomaly detection use cases based on optimized LightGBM and concept drift adaptation. A novel drift adaptation method named Optimized Adaptive and Sliding Windowing (OASW) is proposed to adapt to the pattern changes of online IoT data streams. Experiments on two public datasets show the high accuracy and efficiency of our proposed adaptive LightGBM model compared against other state-of-the-art approaches. The proposed adaptive LightGBM model can perform continuous learning and drift adaptation on IoT data streams without human intervention.
\end{abstract}

\IEEEpeerreviewmaketitle

\section{Introduction}
\begin{figure}
     \centering
     \includegraphics[width=8.5cm]{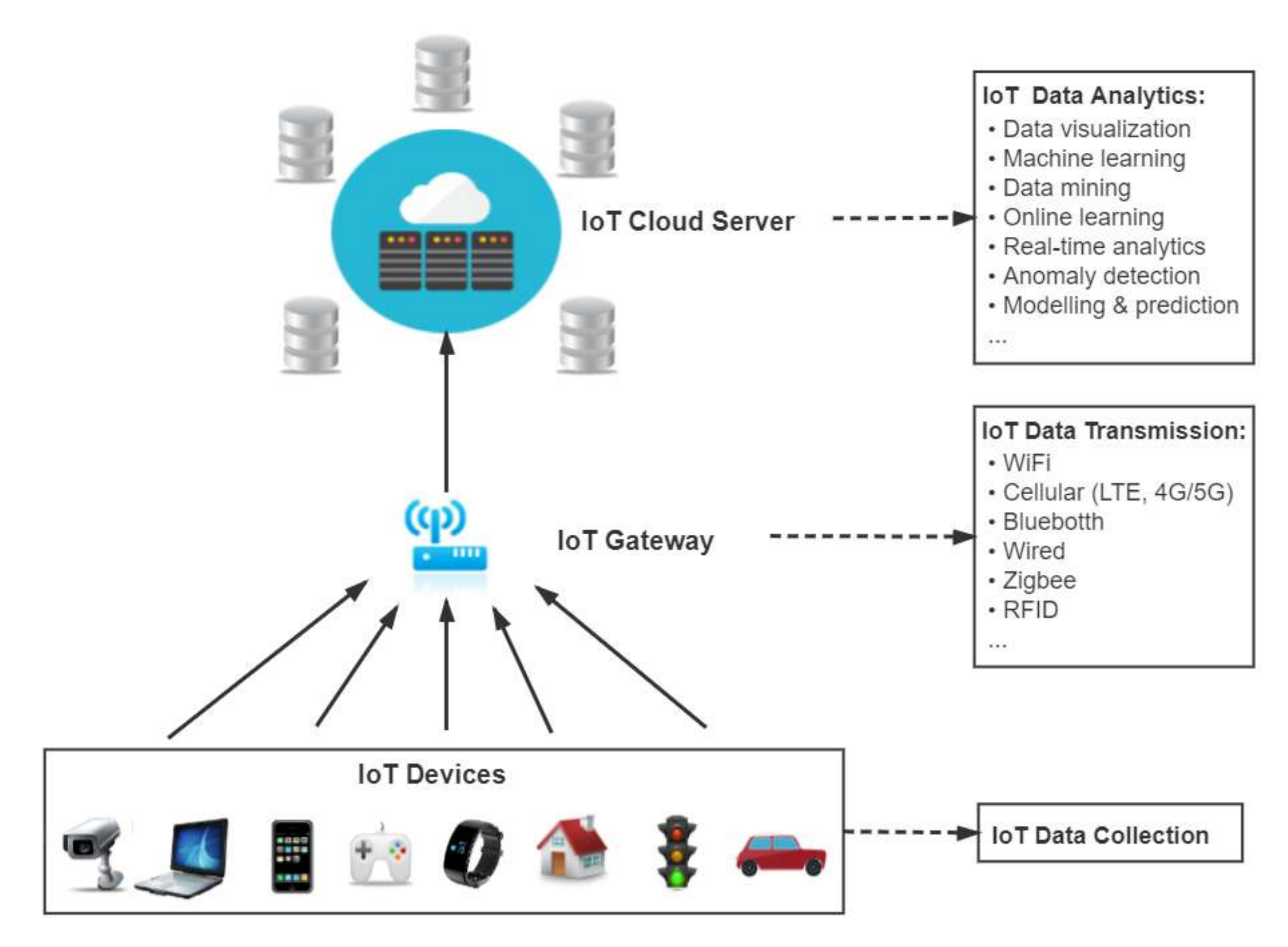}
     \caption{The architecture of IoT data analytics.} \label{iot}
\end{figure}

As Internet and mobile device use grows rapidly, the number of Internet of Things (IoT) devices and the produced IoT data continue to increase significantly. As presented in the Cisco report, more than 30 billion IoT devices are estimated to be connected in 2021, and around five quintillion bytes of IoT data are produced every day \cite{IoT}. IoT data can be collected from different sources and domains, like IoT sensors and smart devices, and can then be transmitted to the central servers for analytics through various communication strategies, including WiFi, Bluetooth, ZigBee, etc. Meaningful insights can be extracted from IoT data streams using data analytics methods to support various IoT applications, like anomaly detection \cite{MAS} \cite{tree}.  Fig. \ref{iot} illustrates the overall architecture of IoT data analytics. 

However, the analytics of tremendous volumes of ever-increasing IoT streaming data and the use of ML models in IoT systems still face many challenges \cite{IoT}. Time and resource constraints are the main issues with IoT data analytics because of the low-power and low-cost requirements of IoT devices \cite{survey}. On the other hand, IoT data is usually dynamic and non-stationary streaming data due to the ever-changing nature of IoT systems and applications. In real-world IoT systems, the physical events monitored by IoT sensors can change over time, and unpredictable abnormal events can occur occasionally. Additionally, IoT device components will age and need update or replacement periodically \cite{causes}. These factors can cause inevitable changes in the statistical distributions of IoT data streams, known as concept drift \cite{survey}. The occurrence of concept drift can cause IoT system failure or performance degradation.
Traditional offline machine learning (ML) models cannot deal with concept drift, making it necessary to develop online adaptive analytics models that can adapt to the predictable and unpredictable changes in the IoT data \cite{drift}.

Therefore, this article proposes a drift adaptive ML-based framework for IoT streaming data analytics. The framework consists of a light gradient boosting machine (LightGBM) for IoT data learning, a particle swarm optimization (PSO) method for model optimization, and a proposed novel method named optimized adaptive and sliding windowing (OASW) for concept drift adaptation. The effectiveness and efficiency of the proposed adaptive framework are evaluated using two public IoT cyber-security datasets: IoTID20 \cite{iotid} and NSL-KDD \cite{NSL}, to evaluate the proposed framework in intrusion detection use cases as an example for IoT data stream analytics.

The main contributions of this article can be summarized as follows: 
\begin{enumerate}
\item We discuss the challenges and potential solutions for IoT streaming data analytics. 
\item We propose a novel drift adaptation method named OASW to address the concept drift issue.\footnote{
code is available at: https://github.com/Western-OC2-Lab/OASW-Concept-Drift-Detection-and-Adaptation} Its performance was evaluated through comparison with other state-of-the-art approaches.
\item We propose an optimized adaptive framework for IoT anomaly detection use cases with offline and online learning functionalities based on LightGBM, PSO, and OASW. 
\end{enumerate}

The remainder of this article is organized as follows: Section II discusses the challenges of IoT data stream analytics. Section III describes the proposed adaptive framework and the novel OASW algorithm. Section IV presents and discusses the experimental results. Section V concludes the article.

\section{IoT Data Analytics Challenges and Potential Solutions}
Due to the high velocity, volume, and variability characteristics of IoT streaming data, there are two major challenges related to IoT data analytics: 1) time \& memory constraints and 2) concept drift \cite{causes}. In this section, the challenges and potential solutions for IoT streaming data analytics are discussed. Table \ref{challenges} summarizes them.

\begin{table*}[ht]
\caption{IoT Data Analytics Challenges and Solutions}
\setlength\extrarowheight{1pt}
\centering
\scalebox{0.98}{
\begin{tabular}{|p{1.75cm}|p{1.4cm}|p{4.7cm}|p{8.5cm}|}
\hline
\textbf{IoT Data Characteristic} & \multicolumn{1}{c|}{\multirow{2}{*}{\textbf{Challenge}}}            & \multicolumn{1}{c|}{\multirow{2}{*}{\textbf{Description}}}                                                                                                                                                                                                                                                                                                           & \multicolumn{1}{c|}{\multirow{2}{*}{\textbf{Potential Solutions}}} \\ \hline

\multirow{10}{*}{\shortstack{High Velocity \\ and Volume}}                    & \multirow{10}{*}{\shortstack{Time \& \\ Memory\\Constraints}}              & Large volumes of IoT data are continuously produced at a high rate, making it difficult to process and store all the data due to the time and memory constraints of low-cost IoT devices. This requires that the average IoT data analytics speed is higher than the average data generation/collection time to meet the real-time processing requirements; otherwise, it could cause IoT service unavailability or system failure.
&  Online learning methods with low computational complexity and forgetting mechanisms are potential solutions to achieve real-time processing of IoT data streams: 
\begin{itemize}
\item[1)] Sliding window methods: They use a sliding window to retain and process only the most recent data samples and discard old samples to save learning time and storage space. 
\item[2)] Incremental learning methods: They can process every incoming new data sample by partially updating the learning model. The data and model complexity can be reduced by discarding the historical data samples and model components to address the execution time and memory constraints of IoT data analytics.    
 \end{itemize}
 \\ \hline

\multirow{14}{*}{High Variability }                & \multirow{6}{*}{\shortstack{Concept \\Drift\\ Detection} }      & Due to the non-stationary IoT data and dynamic IoT environments, concept drift issues often occur in IoT data, causing analytics model degradation. Drift detection faces two main challenges: many causing factors and multiple types of drifts in IoT systems.                                                                                            &   Concept drift can be detected using the following techniques: 
\begin{itemize}
\item[1)] Window-based methods (e.g., ADWIN): 
 They use fixed-sized sliding windows or adaptive windows as data memories for different concepts to detect the occurrence of concept drift. 
\item[2)] Performance-based methods (e.g., DDM \& EDDM): 
 They monitor the model performance degradation rate to detect concept drift.  \end{itemize}         \\ \cline{2-4}
                                 & \multirow{7}{*}{\shortstack{Concept\\ Drift \\Adaptation} }     & After drift detection, the observed drift should be effectively handled so that the learning model can adapt to the new data patterns.                                                                                                                                                                                         &  
Concept drift can be handled using the following techniques: 
\begin{itemize}
\item[1)] Adaptive algorithms (e.g., SAM-KNN): 
They handle concept drift by fully retraining or altering the learning model on an updated dataset after detecting a drift. 
\item[2)] Incremental learning methods (e.g., HATT): 
They partially updated the learning model when new samples arrive or drift is detected. 
\item[3)] Ensemble learning methods (e.g., ARF \& SRP): 
They combine multiple base learners trained on data streams of different concepts. 
 \end{itemize}   \\ \hline

\end{tabular}
}
\label{challenges}%
\end{table*}

\subsection{Time and Memory Constraints}
“High velocity” and “high volume” indicate the high generation speed and large scale of IoT streaming data. This requires that IoT data streams should be processed and analyzed as soon as they reach the learning model. However, in IoT systems, most IoT devices are low-power and low-cost devices with limited computational resources, which limit their data analytics speeds \cite{survey}. The memory constraints of IoT devices also limit their capabilities to process and store large volumes of IoT data and high complexity learning models.  Thus, it is essential to develop low computational complexity analytics models.

Online learning methods that enable real-time analytics are able to satisfy the time and memory constraints of IoT systems. Unlike batch learning techniques that often train a learning model on the entire training set, online learning methods can keep updating the learning model as each new data sample arrives within a short execution time. Sliding window (SW) and incremental learning methods are two potential solutions for IoT online learning \cite{survey}. SW methods retain a limited number of recent data samples and discard the old data samples using sliding windows. Thus, they have forgetting mechanisms that can reduce the storage requirements of IoT devices. Incremental learning is an online learning technique that uses every incoming data sample in the model training and updating process. It can retain the historical patterns and trends of the entire data in the learning model without storing all the data, and adapt to the new data patterns by partially updating the learning model (\textit{e.g.}, replacing the nodes of Hoeffding trees). 

\subsection{Concept Drift Detection}
“High variability” implies the concept drift issue associated with the non-stationary IoT data and the dynamic IoT environments. IoT streaming data is prone to many types of data distribution changes due to the dynamic IoT environments. For example, the physical events monitored by IoT sensors can change over time, and the sensing components age or need updates periodically. The corresponding IoT data distribution changes are named concept drift \cite{causes} \cite{drift}. 

The occurrence of concept drift could degrade the decision-making capabilities of IoT data analytics models, causing severe consequences in IoT systems. For example, the misleading decision-making process of IoT anomaly detection framework could significantly degrade their detection accuracy, making the IoT system vulnerable to various malicious cyber-attacks. To address concept drift, effective methods should be able to detect concept drift and adapt to the changes accordingly. Therefore, concept drift detection and adaptation are the two main challenges related to the high variability issue of IoT streaming data \cite{drift}. 

The first challenge of drift detection is the multiple types of concept drift, including gradual, sudden, and recurring drifts. The second challenge of drift detection is the various factors that can cause concept drift in IoT systems, including the IoT event factors (\textit{e.g.}, system updates, IoT device replacement, and abnormal network events) and time-series factors (\textit{e.g.}, seasonality and trends). 

Window-based methods and performance-based methods are two potential solutions for drift detection \cite{drift}. Adaptive Windowing (ADWIN) is a common window-based method that uses adaptive sliding windows to detect concept drift based on the statistical difference between two adjacent subwindows \cite{drift}. Windowing methods are often fast and easy to implement, but they may lose certain useful historical information.
On the other hand, Drift Detection Method (DDM) and Early Drift Detection Method (EDDM) are two popular performance-based drift detection methods that determine the occurrence of concept drift by monitoring the degree of model performance degradation \cite{drift}. Performance-based methods can effectively detect the drifts that cause model degradation, but they require the availability of ground-truth labels. 

\subsection{Concept Drift Adaptation}

After drift detection, the observed drift should be effectively handled so that the learning model can adapt to the new data patterns. The concept drift adaptation challenge can be addressed using three potential solutions: adaptive algorithms, incremental learning, and ensemble learning methods \cite{drift}.

Adaptive algorithms handle concept drift by fully retraining or altering the learning model on an altered dataset after detecting a drift. They are often the combinations of a ML model and a drift detection technique. For example, Losing \textit{et al.}  \cite{SAMKNN} proposed the self-adjusting memory with k-nearest neighbor (SAM-KNN) algorithm that uses KNN to train a learner and a dual-memory method to store both new and old useful data to fit current and previous concepts.

In incremental learning methods, the learning model is often partially updated when new samples arrive or drift is detected. Manapragada \textit{et al.}  \cite{EFDT} proposed an incremental learning method named Hoeffding Anytime Tree (HATT) that selects and splits nodes as soon as the confidence level is reached instead of identifying the best split in Hoeffding trees. This strategy makes HATT more efficient and accurate to adapt to concept drift. 

Ensemble learning is the technique of combining multiple base learners to construct an ensemble model with better generalization ability. Gomes \textit{et al.}  proposed Adaptive Random Forest (ARF) \cite{ARF} and Streaming Random Patches (SRP) \cite{SRP} ensemble algorithms that both use Hoeffding trees as the base learners and ADWIN as the drift detector. Ensemble models can retain historical and new data patterns in different base learners to address concept drift adaptation, but they often require high execution time.

\section{Proposed System Framework}

\subsection{System Overview}
\begin{figure}
     \centering
     \includegraphics[width=8.8cm]{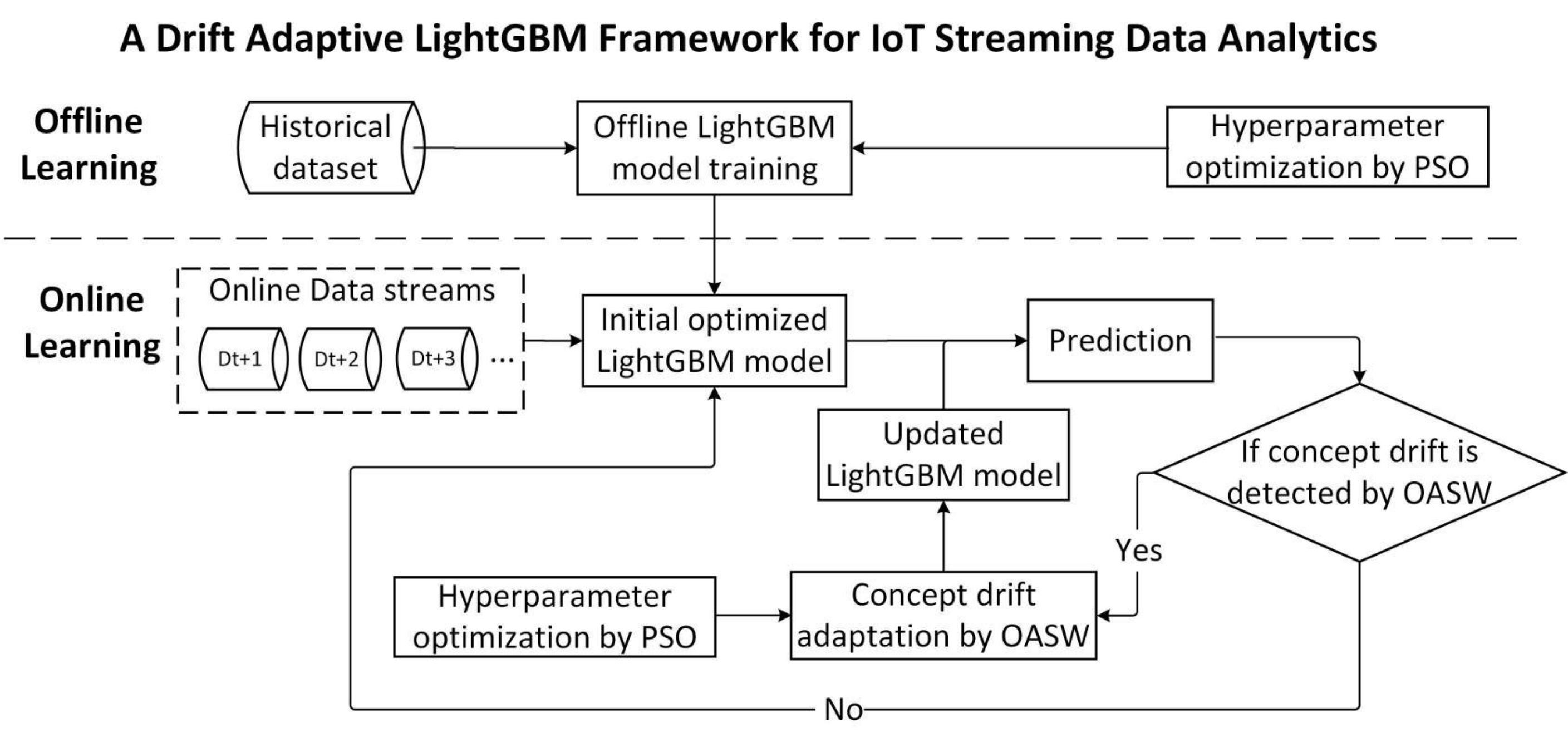}
     \caption{The framework of the proposed drift adaptive IoT data analytics model} \label{frame}
\end{figure}

The purpose of this work is to develop an adaptive IoT streaming data analytics framework that can address the time \& memory constraints, as well as the concept drift issues described in Section II-C. Fig. \ref{frame} demonstrates the overall architecture of the proposed adaptive LightGBM model. It comprises two stages: offline learning to obtain an initial trained model, and online training to detect IoT attacks in online data streams. 

At the offline training stage, current IoT traffic data is collected to create a historical dataset. The historical dataset is then used to train an initial LightGBM model. Moreover, the hyperparameters of the LightGBM model are tuned by PSO, a hyperparameter optimization (HPO) method, to construct the optimized LightGBM model. 

At the online stage, the proposed system will process the data streams that are continuously generated over time. At the beginning of this stage, the initial LightGBM model obtained from the offline learning is used to process the data streams. If concept drift is detected in the new data streams by the proposed OASW method, the LightGBM model will then be retrained on the new concept data samples collected by the adaptive window of OASW to fit the current concept of the new IoT traffic data. Thus, the proposed system can adapt to the ever-changing new IoT traffic data patterns and maintain accurate cyber-attack detection. 

\subsection{Optimized LightGBM}
LightGBM is a fast and high-performance ML model based on the ensemble of multiple decision trees \cite{LightGBM}. Unlike many other ML algorithms, LightGBM is still an efficient model on large-sized and high-dimensional data, mainly due to its two utilized methods: gradient-based one-side sampling (GOSS) and exclusive feature bundling (EFB). GOSS is a down-sampling method that only retains the data samples with large gradients and removes the samples with small gradients during model training, which greatly reduces the time and memory usage. EFB method is used to reduce the feature size by bundling mutually exclusive features, which significantly reduces the training speed without losing important information. 
	
The reasons for choosing LightGBM as the base model for IoT attack detection are as follows:
\begin{enumerate}
\item LightGBM is an ensemble model that has better generalizability and robustness than many other ML algorithms when working on non-linear and high-dimensional data.  
\item By introducing GOSS and EFB, the time and space complexity of LightGBM has greatly reduced from $O(NF)$ to $O(N^\prime B)$, where $N$ and $N^\prime$ are the original and the reduced number of instances, respectively, and $F$ and $B$ are the original and bundled number of features, respectively. 
\item LightGBM has built-in support for categorical data processing and feature selection,  which simplifies the data pre-processing and feature engineering procedures. 
\item LightGBM supports parallel execution by multi-threading, which substantially improves the model efficiency. 
\end{enumerate}

To summarize, LightGBM can achieve both high accuracy and efficiency for data analytics, which is suitable for IoT systems with time and resource constraints.

To build an effective model with high prediction accuracy, the hyperparameters of the LightGBM model are tuned by PSO, a HPO technique. HPO is the process of automatically detecting the optimal hyperparameter values of a ML model to improve model performance \cite{HPO}. 

Among HPO methods, PSO is a popular population-based optimization algorithm that detects the optimal values through communication and cooperation among individuals in a group \cite{HPO}. In PSO, each particle in the swarm will continuously update its own position and velocity based on its current individual best position and the current global best position shared by other particles. Thus, the particles will move towards the candidate global optimal positions to detect the optimal solution.  

The reasons for choosing PSO to tune hyperparameters are as follows \cite{HPO}:
\begin{enumerate}
\item PSO is easy to implement and has a fast convergence speed. 
\item PSO is faster than most other HPO methods, since it has a low computational complexity of $O(N\log N)$ and supports parallel execution.
\item PSO is effective for different types of hyperparameters and large hyperparameter space, like the hyperparameter space of LightGBM. 
\end{enumerate}

Five main hyperparameters, the number of leaves (\textit{num\_leaves}), the maximum tree depth (\textit{max\_depth}), the minimum number of data samples in one leaf (\textit{min\_data\_in\_leaf}), the learning rate(\textit{learning\_rate}), and the number of base learners (\textit{n\_estimators}), are tuned by the PSO method. By detecting the optimal values of these hyperparameters, an optimized LightGBM model can be obtained for accurate IoT data analytics.

\subsection{OASW: Proposed Drift Adaptation Algorithm}

The OASW method is proposed in this article to detect concept drift and adapt to the ever-changing IoT data streams for accurate analytics. OASW is designed based on the combination of ideas in sliding and adaptive window-based methods, as well as in performance-based methods. The complete OASW method is given by Algorithm 1. It has two main functions named “\textit{DriftAdaptation}” and “\textit{HPO}”. The “\textit{DriftAdaptation}” function aims to detect concept drift in the streaming data and update the LightGBM model with the new concept samples for drift adaptation using the given hyperparameter values. The “\textit{HPO}” function is used to tune and optimize the hyperparameters of the “\textit{DriftAdaptation}” function using PSO.

\begin{algorithm}[t]
    {\scriptsize
    %\tiny
	\caption{Optimized Adaptive and Sliding Windowing (OASW)}
	\label{algo:event}
	\LinesNumbered
	\KwIn{
	\\\quad $Stream$: a data stream, 
	\\\quad $\alpha,\beta$: the warning and drift thresholds, 
	\\\quad $t$: the fixed size of a sliding window, \\\quad $t^\prime_{max}$: the maximum size of the adaptive window, 
	\\\quad $Classifier$: a classifier trained on offline dataset, 
	\\\quad $Space$: hyperparameter configuration space, 
	\\\quad $MaxTime$: the maximum hyperparameter search times.}
	\KwOut{
	\\\quad $HP_{opt}$: the detected optimal hyperparameter values, 
	\\\quad $MaxAcc$: the average overall accuracy.}
    \SetKwFunction{FDA}{DriftAdaptation}
    \SetKwFunction{FBO}{HPO}
    \SetKwProg{Fn}{Function}{:}{}
    \Fn{\FDA{$Stream$, $Classifier$, $\alpha$, $\beta$, $t$, $t^\prime_{max}$}}{
	$W^\prime \leftarrow \emptyset$; \tcp*[f]{Initialize the adaptive window}\\
    $State \leftarrow 0$; \tcp*[f]{An indicator of normal, drift, and warning states}\\
    %$c \leftarrow 0$; \tcp*[f]{The current concept drift start point}\\
	\For{\rm{all samples} $x_i\in Stream$}{	
    	$W_i \leftarrow $ a sliding window of the last $t$ samples of $x_i$\; 
	    $AccWin_i \leftarrow accuracy(W_i)$; \tcp*[f]{Current window accuracy}\\
	    $AccWin_{i-t} \leftarrow accuracy(W_{i-t})$; \tcp*[f]{Last window accuracy}\\

	    \If(\tcp*[f]{New window accuracy drops from the normal to warning }){$(Indicator==0)\&\&(AccWin_i<\alpha*AccWin_{i-t})$ } 
    	    {
    	    $W^\prime \leftarrow W^\prime \cup \{x_i\}$ \tcp*[f]{$W^\prime$ starts to collect new samples}\\
    	    $State \leftarrow 1$; \tcp*[f]{Warning occurs}\\
    	    }
	    \If(\tcp*[f]{In a warning state}) {$State==1$}
	    {
	        $t^\prime \leftarrow Size(W^\prime)$\;
	        \uIf(\tcp*[f]{New window accuracy drops to drift level}){$AccWin_i<\beta*AccWin_{i-t}$}	
	        {
    	    $State \leftarrow 2$; \tcp*[f]{Drift occurs}\\
    	    $f \leftarrow i$; \tcp*[f]{Obtain the first new concept window accuracy as a baseline}\\
    	    $Classifier^\prime \leftarrow$ Retrain $Classifier$ on $W^\prime$; \tcp*[f]{Retrain the classifier on new concept samples}\\
    	    }
	 
	    \uElseIf(\tcp*[f]{False alarm (the warning state changes back to normal or stay constant)}) 
	        {$(AccWin_i\ge\alpha*AccWin_{i-t})||t^\prime==t^\prime_{max})$}
	        {
	        $W^\prime \leftarrow \emptyset$; \tcp*[f]{Release the adaptive window}\\
	        $State \leftarrow 0$ \tcp*[f]{Change to a normal state}\\
	        }
	    \Else (\tcp*[f]{Still in the warning state}) {
    	    $W^\prime \leftarrow W^\prime \cup \{x_i\}$ \tcp*[f]{$W^\prime$ keeps collecting new samples}\\
    	    }
        }
        \If(\tcp*[f]{In a drift state}) {$State==2$}
        {
        $t^\prime \leftarrow Size(W^\prime)$\;
        \eIf(\tcp*[f]{When new concept accuracy drops to the warning level or sufficient new concept samples are collected}) {$(AccWin_i<\alpha*AccWin_{f+t})||t^\prime==t^\prime_{max})$}
        {
        $Classifier^\prime \leftarrow$ Retrain $Classifier^\prime$ on $W^\prime$; \tcp*[f]{Construct a robust classifier}\\
        $W^\prime \leftarrow \emptyset$; \tcp*[f]{Release the new concept window}\\
        $State \leftarrow 0$ \tcp*[f]{Change to a normal state}\\
	    }
	    {
    	    $W^\prime \leftarrow W^\prime \cup \{x_i\}$ \tcp*[f]{$W^\prime$ keeps collecting new samples}\\
    	    }
        }
    }
    \KwRet $AvgAcc$; \tcp*[f]{The average accuracy}\\
}

    \SetKwProg{Fn}{Function}{:}{}
    \Fn{\FBO{$Stream$, $Space$, $MaxTime$}}{
        $MaxAcc \leftarrow 0$\;
        \For{$j \leftarrow 1 \  \rm{to} \  \textit{MaxTime}$}
            {	
                $\alpha, \beta, t, t^\prime_{max} \leftarrow SelectConfiguration(Space)$; \tcp*[f]{Search optimal HP values by PSO}\\ 
                $Acc \leftarrow DriftAdaptation(Stream, Classifier, \alpha, \beta, t, t^\prime_{max})$ ; \tcp*[f]{Evaluate the current HP configuration}\\ 
                \If{$MaxAcc<Acc$}{
                $MaxAcc \leftarrow Acc$  ;\\
                $HP_{opt} \leftarrow \alpha, \beta, t, t^\prime_{max}$  ;
                \tcp*[f]{Update accuracy and optimal hyperparameter values}\\
                }
            }
             \KwRet $MaxAcc, HP_{opt}$; \tcp*[f]{The best accuracy \& hyperparameters}\\
        }
}
\end{algorithm}

In OASW, there are two types of windows: a sliding window used to detect concept drift and an adaptive window used to store new concept samples. The size of the sliding window is $t$, and the maximum size of the adaptive window is $t^\prime_{max}$. Additionally, two thresholds, $\alpha$ and $\beta$, are used to indicate the warning level and the actual drift level for concept drift detection. 

The main procedures of OASW are as follows. For each incoming data sample $i$ in the new stream, its sliding window, $W_i$, contains $(i-t)_{th}$ to $i_{th}$ samples. The accuracies of $W_i$ and $W_{i-t}$ (indicating the current and last complete windows, respectively) are calculated and compared. If the accuracy of the sliding window drops $\alpha$ percent from timestamp $i-t$ to $i$, the warning level is reached, and the adaptive window starts to collect incoming data samples as new concept samples (lines 6-12). After that, if the sliding window accuracy keeps dropping $\beta$ percent to the drift level, a drift alarm will be triggered, and the old learner will be updated by retraining on the new concept samples collected in the adaptive window (lines 13-18). 

Moreover, to obtain a robust and stable learner, the adaptive window will keep collecting new samples until one of the following two conditions are met:
1) The new concept accuracy drops to the warning level $\alpha$ when compared to the drift starting point, indicating the current learner is incapable of processing the new concept and requires updating.
2) The size of the adaptive window reaches $t^\prime_{max}$, which ensures that the memory and real-time requirements are met. 

Then, the learner will be updated again on the samples in the adaptive window to become a more robust learner, and the system will change to the normal state for the next potential drift detection (lines 26-35).

On the other hand, if in the warning state, the sliding window accuracy stops dropping, or even increases to the normal level, it will be seen as a false alarm. The adaptive window will then be released, and the system will change back to the normal state to monitor potential new drift (lines 19-24). 

In the OASW algorithm, four parameters, $\alpha$, $\beta$, $t$, and $t^\prime_{max}$, are the critical hyperparameters that have a direct impact on the performance of the OASW model. Therefore, PSO is used to tune these four hyperparameters to obtain the optimized adaptive learner, since PSO is efficient for both continuous and discrete hyperparameters to which the hyperparameters of OASW belong.

To implement OASW, the “\textit{HPO}” function is given the configuration space of the four hyperparameters. PSO will then detect the optimal hyperparameter combination that returns the highest overall accuracy (lines 38-48). The detected optimal hyperparameters will then be given to the “\textit{DriftAdaptation}” function to construct the optimized model for accurate IoT data analytics. 

OASW has a training complexity of $O(NM)$, and a low run-time and space complexity of $O(N)$, where $N$ is the number of instances, and $M$ is the maximum hyperparameter search times in PSO. 

Compared to other concept drift handling methods, the proposed OASW method has the following advantages:
\begin{enumerate}
\item Unlike many other methods that focus on either drift detection or drift adaptation, OASW has both functionalities because it uses a sliding window for drift detection and an adaptive window for drift adaptation.
\item OASW has better generalization capability and adaptability than most other approaches when applied to different datasets or tasks since it can automatically tune the hyperparameters to fit specific datasets by using PSO. 
\item OASW detects concept drift and updates the learning model mainly based on model performance degradation, which ensures that the learner is only updated when necessary. 
\item OASW makes a trade-off between the model accuracy and computational complexity by using the adaptive window to collect sufficient new concept samples while removing previous concept samples. 
\end{enumerate}

\begin{figure*}[t!]
  \centering
  \subfigure[IoTID20.]{
    \label{fig:subfig:a} %% label for first subfigure
    \includegraphics[width=8.7cm]{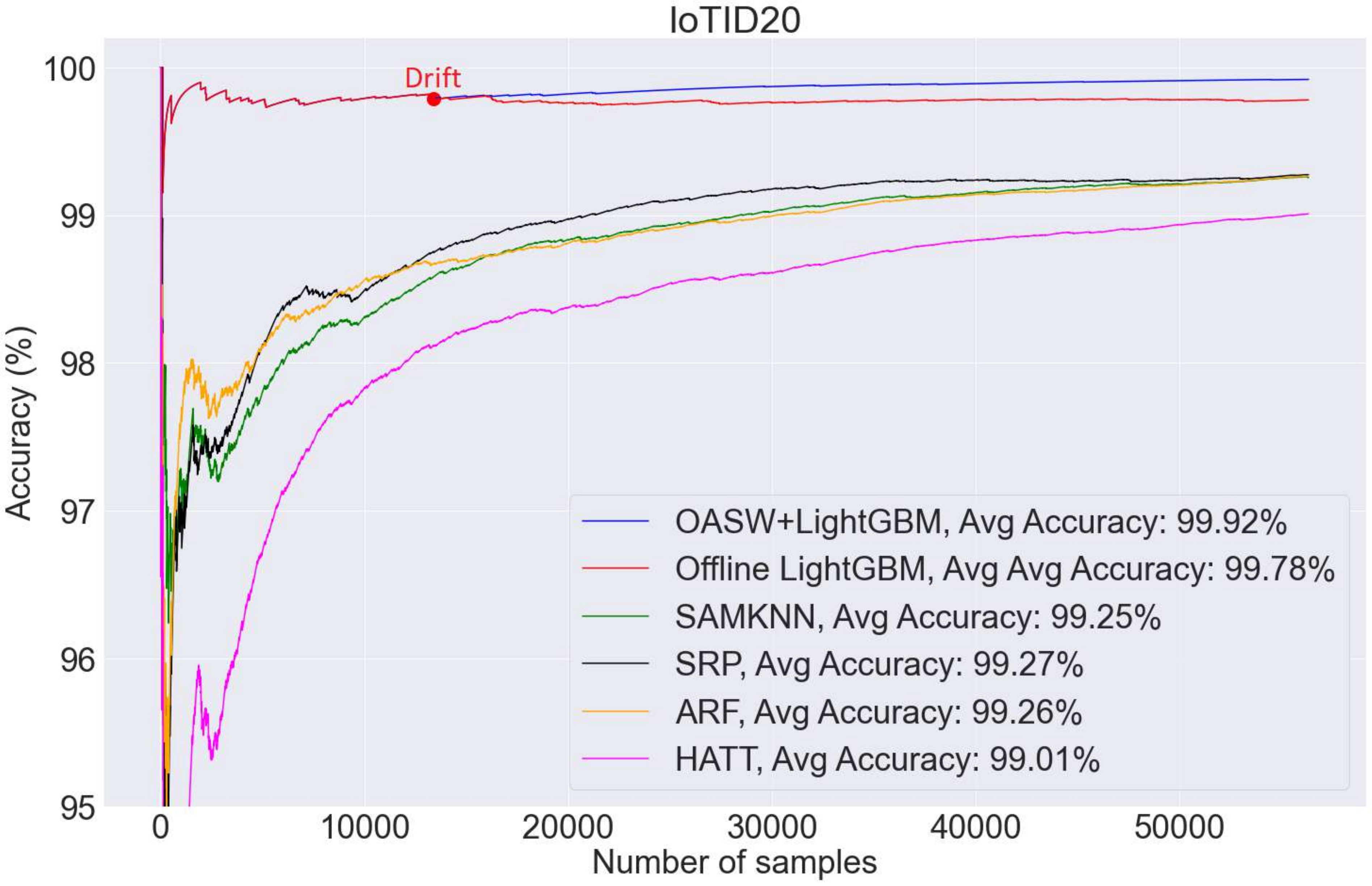}}
  \hspace{0.2cm}
  \subfigure[NSL-KDD.]{
    \label{fig:subfig:b} %% label for second subfigure
    \includegraphics[width=8.7cm]{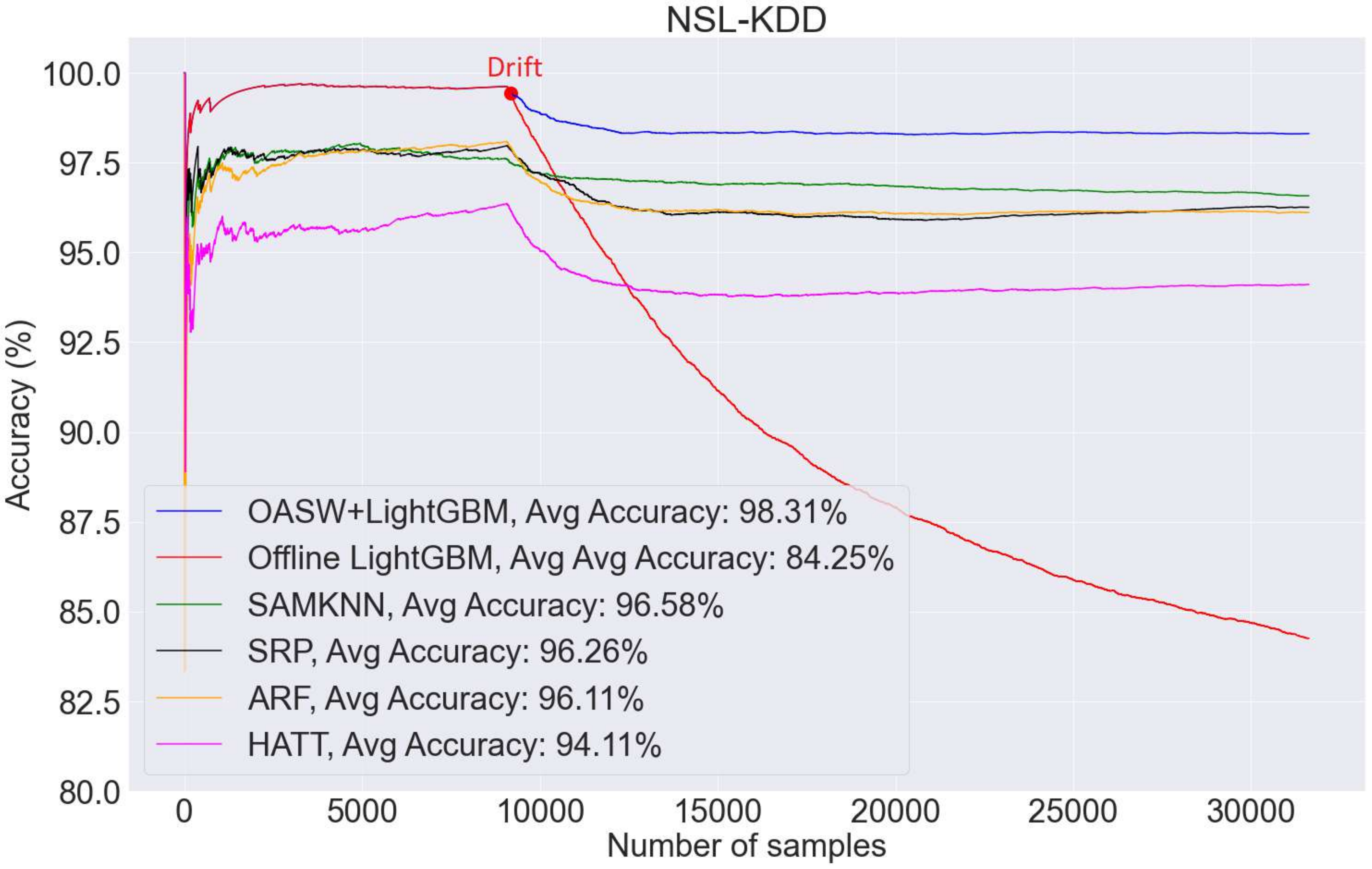}}
  \caption{Accuracy comparison of different drift adaptation methods on two datasets: a) IoTID20; b) NSL-KDD.}
  \label{fig:subfig} %% label for entire figure
\end{figure*}
\section{Performance Evaluation}
\subsection{Experimental Setup}
The experiments were implemented in Python by extending the Scikit-multiflow \cite{skmultiflow} framework on a Raspberry Pi 3 machine with a BCM2837B0 64-bit CPU and 1 GB of memory, representing a real IoT device. Two IoT anomaly detection datasets, IoTID20 \cite{iotid} and NSL-KDD \cite{NSL}, are used to evaluate the proposed framework as examples for IoT classification problems. The proposed framework can be applied to other IoT classification problems using the same procedures. Additionally, the proposed framework can also be applied to general IoT regression problems by only changing the performance metric to a regression metric (\textit{e.g.}, negative mean squared error or negative mean absolute error).

IoTID20 is a novel IoT traffic dataset with unbalanced data samples (94\% normal samples versus 6\% abnormal samples) for abnormal IoT device detection, while NSL-KDD is a balanced benchmark dataset for concept drift and network intrusion detection. They are both widely used in many research projects to validate anomaly detection models in IoT environments. Using these two datasets enables the model evaluation on both balanced and unbalanced datasets. For the purpose of this work, a reduced IoTID20 dataset that has 62,578 records and a reduced NSL-KDD dataset that has 35,140 records are used. The IoTID20 dataset was randomly sampled based on the timestamps (1 data point per 10 timestamps). For the NSL-KDD dataset, it is known that there is a sudden drift from the training set to the test set, but there is no drift in the training set \cite{ARF}; hence, for a clear comparison between the two concepts, the last 10\% of the training set and the entire test set is used for model evaluation.

The proposed method aims to distinguish attack samples from normal samples in IoT systems, so the datasets are utilized as binary classification datasets.  Hold-out and prequential validation methods are used for model evaluation. For hold-out validation, the first 10\% of the data samples in each dataset are used as the training set for offline/initial model training, and the last 90\% are used as the test set for online learning.  Prequential validation, or named test-and-train validation, is only used in online learning, where each input instance is first used to test the model and then used for model updating. To evaluate the performance of the proposed framework, multiple metrics, including accuracy, precision, recall, and f1-score, are used in the experiments. 

For model comparison, four methods introduced in Section II-B, including SAM-KNN \cite{SAMKNN}, HATT \cite{EFDT}, ARF \cite{ARF}, and SRP \cite{SRP}, are also evaluated on the two considered datasets. These four methods are state-of-the-art drift adaptation approaches that have proven effective in many drift datasets and applications. All the drift adaptation methods were implemented using the default parameter values set in Scikit-multiflow or the original papers.

\subsection{Experimental Results and Discussion}
To obtain optimized LightGBM and OASW models, their hyperparameters are automatically tuned by PSO. The initial hyperparameter search range and detected hyperparameter values of the LightGBM and OASW models on the two considered datasets are shown in Table \ref{hps}. After using PSO, the optimal hyperparameter values were assigned to the proposed models to construct optimized models for IoT attack detection.

\begin{table}[]
\centering%
\caption{Hyperparameter Configuration of LightGBM and OASW}
\setlength\extrarowheight{1pt}
\scalebox{0.85}{
\begin{tabular}{|>{\centering\arraybackslash}m{4.5em}|>{\centering\arraybackslash}m{6.8em}|>{\centering\arraybackslash}m{4.8em}|>{\centering\arraybackslash}m{5.5em}|>{\centering\arraybackslash}m{5.5em}|}
\hline
\textbf{Model}            & \textbf{Hyper-parameter} & \textbf{Search Range} & \textbf{Optimal Value (IoTID20)} & \textbf{Optimal Value (NSL-KDD)} \\ \hline
\multirow{5}{*}{LightGBM} & \textit{n\_estimators }           & {[}50, 500{]}         & 300                                      & 300                              \\ \cline{2-5} 
                          & \textit{max\_depth}               & {[}5, 50{]}           & 40                                       & 42                               \\ \cline{2-5} 
                          & \textit{learning\_rate}           & (0, 1)                & 0.56                                     & 0.81                             \\ \cline{2-5} 
                          & \textit{num\_leaves}              & {[}100, 2000{]}       & 200                                     & 100                              \\ \cline{2-5} 
                          & \textit{min\_data\_in\_leaf }     & {[}10, 50{]}          & 35                                       & 45                               \\ \hline
\multirow{4}{*}{OASW}     & $\alpha$                        & (0.95, 1)             & 0.999                                    & 0.978                            \\ \cline{2-5} 
                          &$\beta$                       & (0.90, 1)             & 0.990                                   & 0.954                            \\ \cline{2-5} 
                          & $t$                        & {[}100, 1000{]}        & 300                                      & 350                              \\ \cline{2-5} 
                          & $t^\prime_{max}$                  & {[}500, 5000{]}       & 1000                                     & 3100                             \\ \hline
\end{tabular}
}
\label{hps}
\end{table}

Fig. \ref{fig:subfig} and Table \ref{results} show the accuracy comparison of the proposed OASW \& LightGBM model against the state-of-the-art drift adaptive methods introduced in Section II-C.  It can be seen in Table \ref{results} that the proposed adaptive model outperforms all other methods in terms of accuracy. On the IoTID20, as shown in Fig. \ref{fig:subfig:a}, the proposed method can achieve the highest accuracy of 99.92\% among all implemented models by adapting a slight concept drift detected at point 13408. Without drift adaptation, the offline LightGBM model has a slightly lower accuracy of 99.78\%. The accuracies of the other four state-of-the-art methods are also lower than the accuracy of our proposed method (99.01\% - 99.27\%).

\begin{table*}[]
\centering%
\caption{Performance Comparison of Drift Adaptation Methods}
\setlength\extrarowheight{1pt}
\scalebox{0.85}{
\begin{tabular}{|>{\centering\arraybackslash}m{5.7em}|>{\centering\arraybackslash}m{4em}|>{\centering\arraybackslash}m{4em}|>{\centering\arraybackslash}m{3.5em}|>{\centering\arraybackslash}m{3em}|>{\centering\arraybackslash}m{4.5em}|>{\centering\arraybackslash}m{5.5em}|>{\centering\arraybackslash}m{4em}|>{\centering\arraybackslash}m{4em}|>{\centering\arraybackslash}m{3.5em}|>{\centering\arraybackslash}m{3em}|>{\centering\arraybackslash}m{4.5em}|>{\centering\arraybackslash}m{5.5em}|}
\hline
\multirow{3}{*}{\textbf{Method}} & \multicolumn{6}{c|}{\textbf{IoTID20 Dataset}}                                & \multicolumn{6}{c|}{\textbf{NSL-KDD Dataset}}                                        \\ \cline{2-13} 
                                 & \textbf{Accuracy (\%)}& \textbf{Precision (\%)}& \textbf{Recall (\%)}& \textbf{F1 (\%)} & \textbf{Avg Test Time (ms)} & \textbf{Memory Usage (MB)} & \textbf{Accuracy (\%)} & \textbf{Precision (\%)}& \textbf{Recall (\%)}& \textbf{F1 (\%)} & \textbf{Avg Test Time (ms)} & \textbf{Memory Usage (MB)} \\ \hline
SAM-KNN \cite{SAMKNN}                         & 99.25  & 99.40 & 99.80 & 99.60        & 43.4                         & 179.7                    & 96.58  & 96.13  & 97.6 & 96.86           & 21.6                         & 137.2                      \\ \hline
HATT   \cite{EFDT}                          & 99.01  & 99.21 & 99.74 & 99.47          & 4.8                         & 0.8                       & 94.11  & 94.26 & 94.86  & 94.56         & 3.5                         & 7.3                        \\ \hline
ARF  \cite{ARF}                            & 99.26  & 99.37 & 99.85 & 99.61          & 5.9                         & 0.9                       & 96.11  & 95.81 & 97.05 & 96.42            & 5.6                         & 6.8                        \\ \hline
SRP \cite{SRP}                             & 99.27  & 99.35 & 99.88 & 99.61           & 67.8                        & 3.8                        & 96.26    & 96.21 & 96.88 & 96.55         & 35.7                         & 15.3                       \\ \hline
Offline LightGBM                 & 99.78  & 99.82 & 99.95 & 99.88          & 0.6                        & 1.9                       & 84.25    & 97.75 & 72.51 & 83.26        & 0.3                        & 3.7                        \\ \hline
Proposed OASW \& LightGBM                  & 99.92  & 99.93 & 99.98 & 99.96         & 7.8                         & 0.4                       & 98.31   & 98.57 & 98.30   & 98.43        & 9.1                         & 1.8                        \\ \hline
\end{tabular}
}
\label{results}
\end{table*}

For the NSL-KDD dataset, there is a severe drift at the beginning of the test set \cite{ARF}. As shown in Fig. \ref{fig:subfig:b} and Table \ref{results}, by adapting to the sudden drift detected at point 9183, the proposed method can achieve the highest accuracy of 98.31\%, while the offline LightGBM model’s accuracy drops significantly to only 84.25\% without drift adaptation. This emphasizes the development of drift adaptation methods. The other four compared methods, SAM-KNN, SRP, ARF, and HATT, have much lower accuracy than the proposed method (94.11\% - 96.58\%). 

The average online prediction time for each instance and the total memory usage are also calculated to evaluate the proposed method for real-time online learning considering the time and memory limitations of IoT devices, as shown in Table \ref{results}. Among the six methods implemented, the memory usage of the proposed method is the smallest on both datasets, because the proposed model updates itself on a relatively small subset obtained by OASW instead of on the entire streaming data. The average prediction time of the proposed model for each instance on the Raspberry Pi 3 machine is only 7.8 ms and 9.1 ms on the two used datasets, much shorter than SAM-KNN and SRP. This is mainly due to the sliding window strategy and the efficiency of LightGBM. The prediction time of ARF and HATT for each instance is shorter than the proposed model, but their accuracy is much lower than the proposed method. Therefore, the proposed method still performs the best among the drift methods in terms of the trade-off between accuracy and efficiency. Moreover, the average execution time of the proposed framework for each sample on a desktop machine with an i7-8700 processor \& 16 GB of memory and a powerful Google Colaboratory Cloud machine with a Xeon processor is only 1.3 ms and 0.9 ms, respectively. The high data processing speed on powerful cloud machines shows the feasibility of implementing the proposed framework in real-time environments. 

In conclusion, the experimental results show the effectiveness and robustness of the proposed adaptive LightGBM model for IoT streaming data analytics.

\section{Conclusion}
The increasing popularity of IoT systems has brought great convenience to humans, but it also increases the difficulty to collect and process large volumes of IoT data collected from various sensors in IoT environments. Compared to conventional static data, IoT data is often big streaming data under non-stationary and rapidly-changing environments. Adaptive ML methods are appropriate solutions since they have the capacity to process constantly evolving IoT data streams by adapting to potential concept drifts. In this article, we proposed the adaptive LightGBM model for IoT data analytics with high accuracy and low time and memory usage. Through the integration of our proposed novel drift-handling algorithm (\textit{i.e.}, OASW), an ensemble ML algorithm (\textit{i.e.}, LightGBM), and a hyperparameter method (\textit{i.e.}, PSO), the proposed model has the capacity to automatically adapt to the ever-changing data streams of dynamic IoT systems. The proposed method is evaluated and discussed by conducting experiments on two public IoT anomaly detection datasets, IoTID20 and NSL-KDD. Based on the comparison with several state-of-the-art drift adaptation methods, the proposed system is able to detect IoT attacks and adapt to concept drift with higher accuracies of 99.92\% and 98.31\% than the other methods on the IoTID20 and NSL-KDD datasets, respectively.


%\section*{Acknowledgment}

%\balance
\begin{IEEEbiography}[{\includegraphics[width=1in,height=1.25in,clip,keepaspectratio]{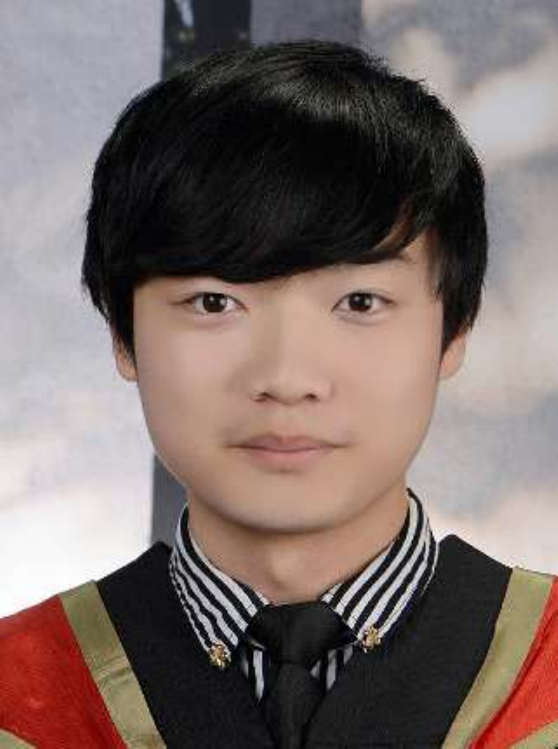}}]{Li Yang} [M] (lyang339@uwo.ca) received the B.E. degree in computer science from Wuhan University of Science and Technology, Wuhan, China in 2016 and the MASc degree in Engineering from University of Guelph, Guelph, Canada, 2018. Since 2018 he has been working toward the Ph.D. degree in the Department of Electrical and Computer Engineering, Western University, London, Canada. His research interests include cybersecurity, machine learning, time-series analytics, and IoT data analytics.
\end{IEEEbiography}

\begin{IEEEbiography}[{\includegraphics[width=1in,height=1.25in,clip,keepaspectratio]{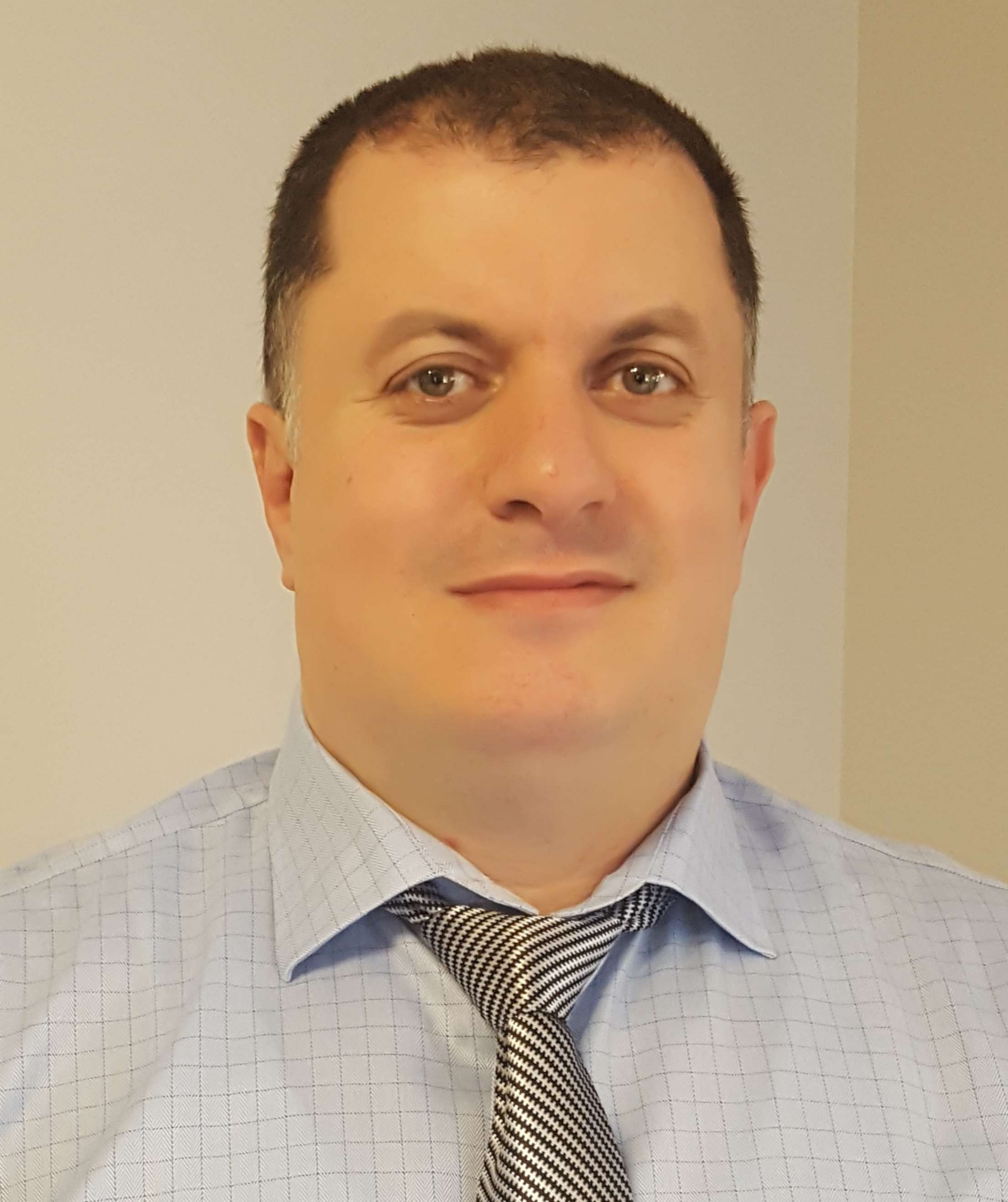}}] {Abdallah Shami} [SM] (abdallah.shami@uwo.ca) is a professor with the ECE Department at Western University, Ontario, Canada. He is the Director of the Optimized Computing and Communications Laboratory at Western University (https://www.eng.uwo.ca/oc2/). He is currently an associate editor for IEEE Transactions on Mobile Computing, IEEE Network, and IEEE Communications Surveys and Tutorials. He has chaired key symposia for IEEE GLOBECOM, IEEE ICC, IEEE ICNC, and ICCIT. He was the elected Chair of the IEEE Communications Society Technical Committee on Communications Software (2016-2017) and the IEEE London Ontario Section Chair (2016-2018).
\end{IEEEbiography}

\end{document}